\documentclass[runningheads]{llncs}
\usepackage[T1]{fontenc}
\usepackage{amsmath}
\usepackage{amssymb}
\usepackage{color}
\usepackage{graphicx}
\usepackage[
    bookmarks=true,
    colorlinks=true,
    linkcolor=blue,
    urlcolor=blue,
    citecolor=blue,
    bookmarks=true,
    linktocpage=true,
    hyperindex=true
]{hyperref}
\usepackage[misc]{ifsym}
\usepackage{hyperref}

\begin{document}

\title{HCNQA: Enhancing 3D VQA with Hierarchical Concentration Narrowing Supervision}

\titlerunning{Enhancing 3D VQA with Hierarchical Concentration Narrowing Supervision}

\author{Shengli Zhou\inst{1}\and
Jianuo Zhu\inst{1}\and
Qilin Huang\inst{1}\and
Fangjing Wang\inst{1}\and
Yanfu Zhang\inst{2}\and
Feng Zheng\inst{1,3}(\Letter)
}

\authorrunning{S. Zhou et al.}

\institute{
Southern University of Science and Technology, Shenzhen, China
\and
College of William and Mary, Williamsburg, VA 23185, USA
\and
Peng Cheng Laboratory, Shenzhen, China
\\
\email{f.zheng@ieee.org}
}
\maketitle

\begin{abstract}

3D Visual Question-Answering (3D VQA) is pivotal for models to perceive the physical world and perform spatial reasoning. Answer-centric supervision is a commonly used training method for 3D VQA models. Many models that utilize this strategy have achieved promising results in 3D VQA tasks. However, the answer-centric approach only supervises the final output of models and allows models to develop reasoning pathways freely. The absence of supervision on the reasoning pathway enables the potential for developing superficial shortcuts through common patterns in question-answer pairs. Moreover, although slow-thinking methods advance large language models, they suffer from underthinking. To address these issues, we propose \textbf{HCNQA}, a 3D VQA model leveraging a hierarchical concentration narrowing supervision method. By mimicking the human process of gradually focusing from a broad area to specific objects while searching for answers, our method guides the model to perform three phases of concentration narrowing through hierarchical supervision. By supervising key checkpoints on a general reasoning pathway, our method can ensure the development of a rational and effective reasoning pathway. Extensive experimental results demonstrate that our method can effectively ensure that the model develops a rational reasoning pathway and performs better. The code is available at \url{https://github.com/JianuoZhu/HCNQA}.

\keywords{Human-like reasoning \and Reasoning pathway regularization \and Shortcut suppression \and 3D Visual Question-Answering.}
\end{abstract}

\section{Introduction}

3D Visual Question-Answering (3D VQA) involves answering questions based on given 3D scenarios. It can significantly enhance models' ability for 3D environment perception and spatial reasoning through natural language.

Since common answers to a specific question are limited, models may easily guess the answer without performing spatial reasoning. For example, when asking about the shape of a table next to the door, the model can answer ``rectangle'' without knowing which table the question is about. Thus, when training 3D VQA models, a common method for optimizing models is to supervise answer-centric information (i.e., supervising outputs after all inference steps) beyond textual answers. For instance, ScanQA \cite{ScanQA}, LL3DA \cite{LL3DA}, and BridgeQA \cite{BridgeQA} require object localization using bounding boxes, 3D-VisTA \cite{3DVisTA} outputs the indices of target objects,\footnote{The target object corresponding to a question is the object that directly answers the question or the object that contains some attribute as the answer to the question.} while SQA3D \cite{SQA3D} predicts positions and directions for the agent when solving the questions. These models, using the aforementioned supervision methods, ensure that they refer to the correct object when answering questions. As answer-centric supervision methods have proven beneficial for improving performance, they have become a mainstream training approach for 3D VQA models.

\begin{figure}[t]
    \centering
    \includegraphics[width=0.9\textwidth]{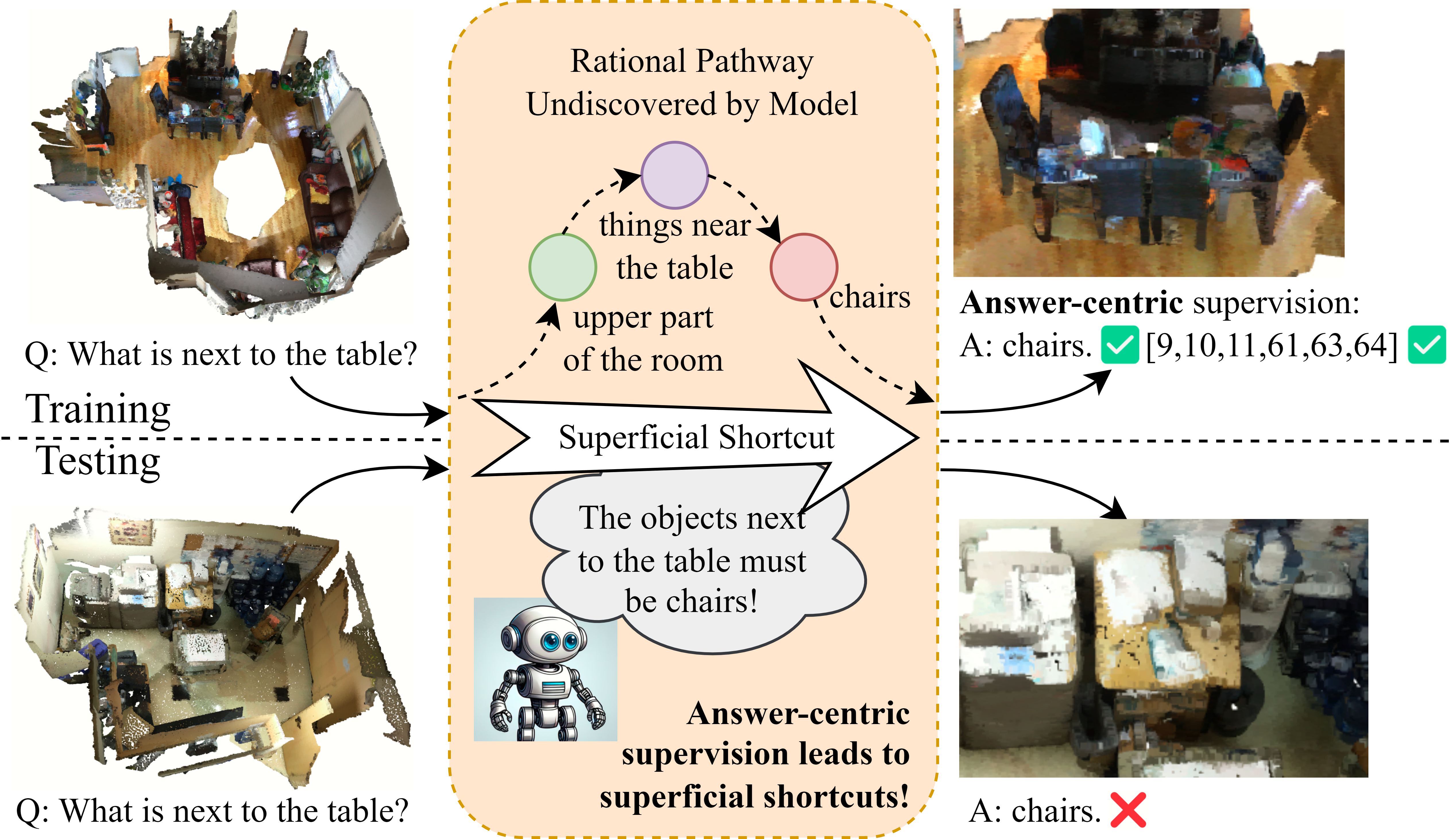}
    \caption{In answer-centric approaches, the absence of supervision on the reasoning pathway can lead models to adopt shortcuts, such as predicting answers before identifying relevant objects. As shown in the figure, during training, the model formed an incorrect shortcut between table and chairs, not adopting the correct reasoning pathway marked with dashed lines. This faulty correlation ultimately led to a failure during testing.}
    \label{fig:demo}
\end{figure}

Although these methods guide the model in referencing the correct object, they only supervise the final outcome after all inference steps. This leaves room for models to develop their own reasoning strategies, as demonstrated in research \cite{shortcut_1,shortcut_2}, which may involve shortcuts (as depicted in Fig. \ref{fig:demo}). As defined in \cite{shortcut_1}, shortcuts occur when ``\textit{models have many opportunities to bypass reasoning and instead find shallow patterns in the data to match answers to scene-question pairs.}'' When models are applied to unseen instances, these shortcuts can easily mislead them, resulting in unsatisfactory answers.

Meanwhile, although models like OpenAI o1 \cite{gpto1} and DeepSeek-R1 \cite{deepseekr1} have gained improvements from slow-thinking techniques, they suffer from the issue of underthinking. \cite{underthinking} That is, these models may shift frequently between multiple reasoning pathways without making rapid progress. Therefore, a supervision method for regulating the reasoning pathway with high reasoning efficiency and rationality is required to achieve a better balance.

To address these issues, we propose \textbf{HCNQA}, a  \underline Hierarchical \underline Concentration \underline Narrowing Supervision-based \underline Question-\underline Answering model. In HCNQA, rather than supervising the result of each intermediate step, we divide the reasoning process into three major phases. By supervising the result of each phase, we guide the model to develop a rational reasoning pathway, thereby suppressing the formation of shortcuts while ensuring effective progress during reasoning. Inspired by human spatial reasoning \cite{cognition}, we propose a general 3D VQA reasoning pathway in which the model progressively narrows the concentration region until the answer is found. Specifically, we leverage a hierarchical supervision module in HCNQA that filters objects in the scene through three phases of object-level mask filtering sequentially (namely, coarse grounding, fine grounding, and inference). The phases in the proposed module respectively obtain \underline Blocks \underline of \underline Interest (BoI), \underline Objects \underline of \underline Interest (OoI) and \underline Object \underline of \underline Target (OoT), as detailed in Sec. \ref{sec:hcnrp}. By supervising the predictions of intermediate steps, we can ensure that the model indeed performs spatial reasoning before giving the answer, thus suppressing the development of superficial shortcuts and improving performance.

In summary, our primary contributions are:

\begin{itemize}
    \item We propose HCNQA, a model that regulates the reasoning pathway through hierarchical concentration narrowing supervision of key intermediate results, suppressing superficial shortcuts, and improving the model's performance.
    \item We establish the first structured reasoning pathway paradigm - a human-inspired three-phase spatial reasoning framework designed for 3D VQA, which effectively ensures the reasoning progress during inference.
    \item We validate the effectiveness of hierarchical concentration narrowing supervision through comparative experiments and ablation studies. Extensive results demonstrate the improvements after applying our methods.
\end{itemize}

\section{Related Work}

\subsection{3D Visual Question-Answering}

The field of 3D Visual Question-Answering (3D VQA) represents a significant leap in the intersection of computer vision and natural language processing. In 3D VQA, models are designed to answer questions based on 3D scenarios. Generally, 3D VQA models can be divided into classification- and generation-based models. Both types of models employ a language encoder (e.g., BERT \cite{BERT}) and a vision encoder (e.g., PointNet \cite{pointnet}) to extract features from mono-modal input, and a vision-language fusion layer to integrate the extracted features. Classification-based models \cite{ScanQA,MCLIP,3DVisTA} consider the 3D VQA task as choosing words from a fixed corpus. For generation-based models \cite{LL3DA,3DLLM,LEO}, they employ Large Language Model (e.g., \cite{flan}) or pretrained Vision-Language Model (e.g., \cite{clip}) with a language decoder to align features and generate answers.

\subsection{Auxiliary Information Supervision}

In the field of 3D vision-language research, the integration of auxiliary information supervision has become a common strategy to steer models away from superficial reasoning pathways. Models such as 3D JCG \cite{3DJCG} incorporate various auxiliary losses, including object detection (DET), object classification (O-CLS), and text classification (T-CLS). Similarly, ViL3DRef \cite{ViL3DRef} leverages knowledge distillation, O-CLS, and T-CLS. MVT \cite{MVT} utilizes O-CLS and T-CLS, while ScanQA \cite{ScanQA} uses DET and O-CLS to enhance models' spatial understanding. Pretrained models like 3D-VisTA \cite{3DVisTA} mainly utilize task loss (e.g., sentence loss and object localization loss for 3D VQA) when fine-tuning downstream tasks.

Currently, supervised information in 3D VQA is primarily answer-centric. For instance, models such as ScanQA \cite{ScanQA}, CLIP-Guided \cite{CLIPG}, Multi-CLIP \cite{MCLIP}, and BridgeQA \cite{BridgeQA} apply supervision on bounding boxes of target objects, classes of target objects, and textual answers. Models such as 3D-ViSTA \cite{3DVisTA}, which utilize features based on segmented scenes, require the model to predict a set of objects that corresponds to the targets. Generation-based models, such as 3D-LLM \cite{3DLLM}, LL3DA \cite{LL3DA} and LEO \cite{LEO} have also required the model to predict bounding boxes for target localization. These models typically apply textual loss to all generated tokens, including answer and auxiliary information.

Despite recent advancements, answer-centric supervision overlooks the intermediate steps of reasoning. As answer-centric supervision allows the model to develop any intermediate reasoning pathway, the model can easily generate superficial shortcuts based on common patterns in question-answer pairs. Additionally, these practices may hinder the model's ability to fully comprehend spatial relationships between objects, deteriorating the quality of learning.

\section{Method}

In this section, we propose a hierarchical concentration narrowing framework to suppress superficial shortcuts while enhancing the interpretability of the reasoning process. Inspired by how humans perform spatial reasoning, the framework divides the reasoning pathway into three phases and enforces the model to narrow down its area of concentration in each phase. The outcome of each phase is supervised to guide the progression of the reasoning pathway. In the architecture of HCNQA, we integrate a pre-HSM feature extractor to enhance visual features for spatial reasoning. Then, we employ a hierarchical supervision module (HSM) to guide the model in explicitly learning the correct sequence of the reasoning process, thereby preventing shortcuts that guess intermediate steps by subsequent results. Finally, we apply the weighted cross-entropy loss to address the class imbalance when evaluating predictions of object-level masks.

\subsection{Hierarchical Concentration Narrowing Reasoning Pathway} \label{sec:hcnrp}

Due to underthinking, it is suboptimal for the model to generate a detailed question analysis through slow-thinking techniques. Thus, to guarantee the model's progress through intermediate-step supervision, a simple and general reasoning pathway that is independent of the given question should be proposed.

To develop such a pathway, we analyze a multitude of question-answer pairs and formulate a reasoning pathway by mimicking humans' way of solving the inquiries. The pathway starts from viewing the full scene and generally narrows down the model's concentration area until it reaches the answer. Furthermore, to supervise the result of each phase, we generate pseudo-labels for ground truths by applying heuristic rules on ScanQA's annotation (which itself does not have a clear meaning). The generated labels are in the form of object-level masks (i.e., a set of selected objects). The details of each phase are as follows:

\begin{enumerate}
    \item \textbf{Block of Interest (BoI)} Initially, survey the environment to ascertain the direction or room segment that the model should focus on. Inspired by the YOLO \cite{YOLOv1} framework, we divide the scene into a $S\times S$ grid, with each cell representing an equal-sized portion of the scene based on $x$ and $y$ coordinates. A cell is deemed a Block of Interest (BoI) if it contains any points from the target or anchors\footnote{We define anchors as the objects mentioned in the question, which act as the role to assist the identification of the target object.}. All objects that exist in the blocks of interest are selected in the object-level mask corresponding to BoI.
    \item \textbf{Objects of Interest (OoI)} Subsequently, determine the Region of Interest, which is defined as the objects that are either the target or anchors.
    \item \textbf{Object of Target (OoT)} Finally, identify the objects that are directly involved in formulating a response, i.e., the target to the question.
\end{enumerate}

By supervising the model to sequentially generate BoI, OoI, and OoT, the reasoning pathway can be regulated (narrowing the concentration region from part of the room to target objects), and shortcuts can be suppressed, thereby enhancing the rationality of the reasoning steps.

\begin{figure*}[t]
    \centering
    \includegraphics[width=0.9\textwidth]{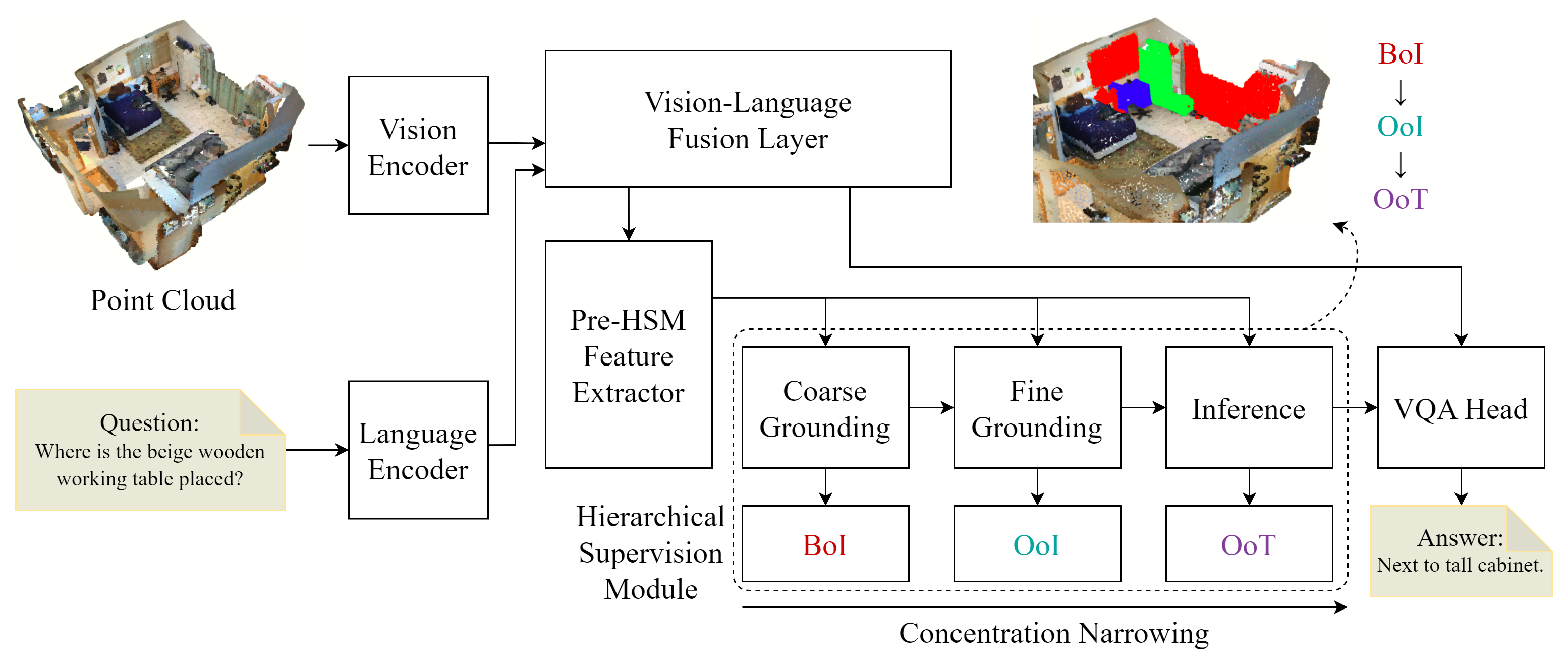}
    \caption{The architecture of the HCNQA model. The Hierarchical Supervision Module (HSM) is boxed out in the model. It includes the coarse grounding, fine grounding, and inference modules. The modules in HSM generate the results of hierarchical concentration narrowing (i.e., BoI, OoI, and OoT), respectively. The information from each module also serves as a reference for the subsequent module. A VQA head is placed after the inference module to generate textual answers based on the results of hierarchical concentration narrowing and the fused vision-language features.}
    \label{fig:hcnqa}
\end{figure*}

\subsection{Model Architecture}

Our model's pipeline consists of feature encoders and an extractor, a Hierarchical Supervision Module (HSM) to supervise the reasoning pathway, and a VQA head to generate textual responses. The architecture of our model is shown in Fig. \ref{fig:hcnqa}.

\subsubsection{Feature Encoder} Following previous works in 3D-VQA, we adopt a common structure for encoding and fusing features. That is, we employ a language encoder to encode the textual question, a vision encoder to encode the point cloud of the environment, and a vision-language fusion layer to fuse the encoded features.

\subsubsection{Pre-HSM Feature Extractor} After feature fusion, the resulting feature is represented as a series of object tokens $\overrightarrow o=[o_1, o_2, \cdots, o_n]^T$, where $o_i$ is the feature of the $i$-th object. In order to further extract features of the entire scene to support spatial reasoning better, we apply a four-layer MLP to transform the object tokens into a base 3D feature, denoted as $F_\text{base}$.

\subsubsection{HSM} 
\begin{figure*}[t]
    \centering
    \includegraphics[width=0.9\textwidth]{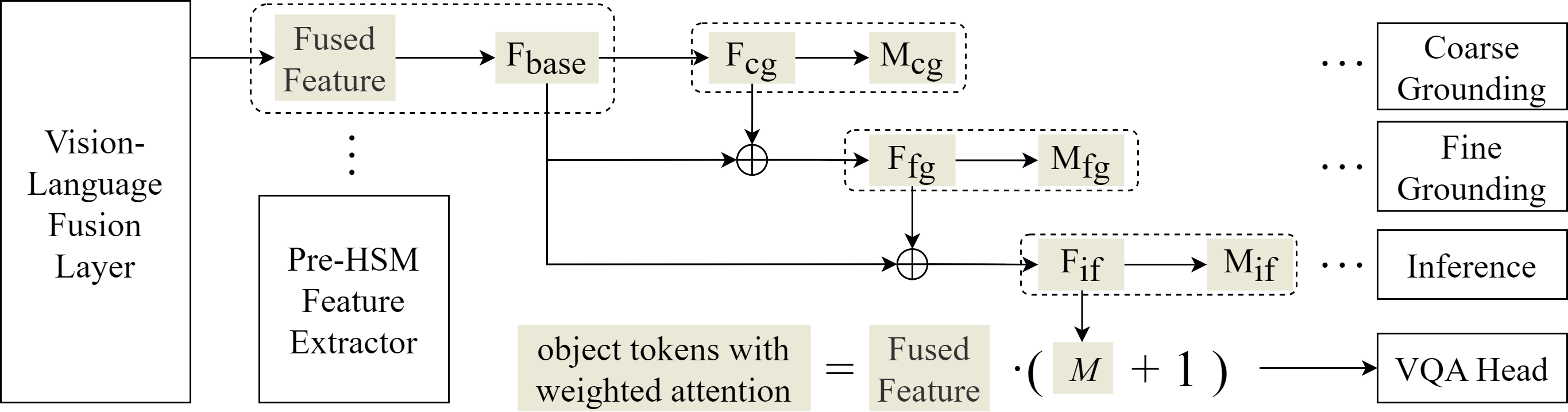}
    \caption{The architecture of the Hierarchical Supervision Module (HSM). The arrows in the diagram indicate the direction of data flow throughout the processing stages.}
    \label{fig:hsm}
\end{figure*}

The Hierarchical Supervision Module (HSM) consists of $3$ submodules, namely coarse grounding, fine grounding, and inference. These modules correspond to the reasoning phases in Section \ref{sec:hcnrp}. HSM is designed as a sequential structure to ensure that the model performs reasoning step by step instead of guessing out the answer and making up the intermediate steps later. The module starts with $F_\text{base}$ and the submodules predicts $F_\text{cg}$, $F_\text{fg}$, and $F_\text{if}$ respectively. The feature extracted from the previous module is concatenated with $F_\text{base}$ and serves as the input of the subsequent module. The features extracted from each submodule are then transformed into the predictions of object-level masks, namely $M_\text{cg}$, $M_\text{fg}$, and $M_\text{if}$. The overall structure for HSM is illustrated in Figure \ref{fig:hsm}.

\subsubsection{VQA Head}

To leverage the information predicted for targets, the VQA head first transforms OoT masks into a vector $\overrightarrow M=[M_1, M_2, \cdots, M_n]^T$, where $M_i$ is the adjustment weight for object $i$. Then, each object token $o_i$ is multiplied by a factor of $M_i+1$ to obtain object tokens with weighted attention (denoted as $o_i'$). The constant $1$ is introduced to avoid ignorance of objects that are not selected by OoT and enhance fault tolerance.

Finally, we feed text tokens from the vision-language fusion layer and $\{o_i'\}$ from HSM into a Modular Co-Attention Network (MCAN) \cite{MCAN} to generate textual answers to questions.

The lightweight nature of HSM's MLP architecture results in minimal supervision costs, with merely $7.30\%$ additional FLOPs introduced during each forward propagation.

\subsection{Supervision Loss for HSM}

Since the predictions of object-level masks are represented as a probability for each class, cross-entropy loss can be applied to optimize the model. However, the objects selected in BoI, OoI, and OoT only account for a minor portion of all objects in the scene. For example, only $1.709$ objects per question are selected as OoT on average. This leads to a significant class imbalance when training the model to predict object-level masks as a binary classification problem.

To mitigate this issue, we apply weighted binary cross-entropy loss when calculating the object localization loss for each phase:

\begin{equation}
    \mathcal L_\text{obj}=-\dfrac{c_0+c_1}N\sum\limits_{i=1}^N\left[\dfrac{M_i}{c_1}\log(\hat{M_i})+\dfrac{1-M_i}{c_0}\log(1-\hat{M_i})\right]
\end{equation}

\noindent where $c_0, c_1$ are the number of unselected and selected objects, respectively.

By applying this method, we can calculate $\mathcal L_\text{cg}, \mathcal L_\text{fg}$ and $\mathcal L_\text{if}$. Finally, the total loss for HSM is defined as: $\mathcal L_\text{HSM} = \lambda_\text{cg}\mathcal L_\text{cg} + \lambda_\text{fg}\mathcal L_\text{fg} + \lambda_\text{if}\mathcal L_\text{if}$.

\section{Experiments}

\subsection{Experimental Settings}

\subsubsection{Implementation Details}

For BoI, we segment the scene into a $5 \times 5$ grid (i.e., $S = 5$). When calculating $\mathcal L_\text{HSM}$, we set $\lambda_\text{cg} = 0.2, \lambda_\text{fg} = 0.3$ and $\lambda_\text{if} = 0.5$. Since the object-level masks are based on object segmentation, we use ground-truth segmentation or, if unavailable, predictions from Mask3D \cite{Mask3D} to ensure consistency of object segmentation across experiments.

During training, we first follow 3D-VisTA \cite{3DVisTA} by pretraining our model's encoders and vision-language fusion layer on Masked Language Modeling, Masked Object Modeling, and Scene-Text Matching tasks. Subsequently, we train the entire model using hierarchical concentration narrowing supervision on a single NVIDIA Titan RTX 24GB GPU.

\subsubsection{Evaluation Metrics}

To evaluate the model's performance, we calculate the accuracy of the answers with the highest confidence (EM@1) and the answers with the top 10 highest confidence (EM@10). Additionally, we apply common sentence evaluation metrics, including BLEU, ROUGE-L, Meteor, and CIDEr.

\begin{table*}[t]
    \centering
    \caption{Results of Annotation Quality Verification on ScanQA's Validation Split}
    \begin{tabular}{c|cccccccccc}
    \hline
        Mask & EM@1 & $\gamma$ & EM@10 & $\gamma$ & BLEU-4 & $\gamma$ & ROUGE-L & $\gamma$ & CIDEr & $\gamma$  \\ \hline
        None & 22.05  & 0.00  & 50.31  & 0.00  & 10.12  & 0.00  & 33.33  & 0.00  & 63.17  & 0.00   \\ 
        \verb|object_ids| & 22.20  & 1.00  & 52.04  & 1.00  & 12.71  & 1.00  & 35.00  & 1.00  & 68.24  & 1.00   \\ 
        BoI & 22.65  & 4.00  & 52.43  & 1.23  & 14.41  & 1.66  & 34.60  & 0.76  & 67.72  & 0.90   \\ 
        OoI & 23.72  & 11.13  & 52.15  & 1.06  & 10.33  & 0.08  & 35.93  & 1.56  & 68.66  & 1.08   \\ 
        OoT & 23.25  & 8.00  & 52.36  & 1.18  & 12.30  & 0.84  & 35.26  & 1.16  & 67.27  & 0.81  \\ \hline
    \end{tabular}
    \label{tab:aqv}
\end{table*}

\subsection{Annotation Quality Verification} \label{sec:aqv}

To verify the effectiveness of BoI, OoI, and OoT annotations, we conduct experiments by training models without and with object localization supervision using either ScanQA's \verb|object_ids| annotations, BoI, OoI, or OoT. The results are shown in Table \ref{tab:aqv}. We compute the improvement ratio $\gamma$ between the score improvements using our annotations and those from ScanQA.

The results illustrate that supervision using \verb|object_ids|, BoI, OoI, and OoT all positively impact model performance. The improvements indicate that supervising the target or anchors located by the model can eliminate the cases where the model guesses out the answer without performing spatial reasoning. Moreover, models trained under supervision from BoI, OoI, and OoT generally perform better. The improvements demonstrate that these annotations have higher quality and are easier for the model to learn and utilize.

\begin{table*}[t]
    \centering
    \caption{Results of Ablation Study on ScanQA's ``Test w/o obj'' Split}
    \begin{tabular}{cccc|cccccccc}
    \hline
        \multicolumn{4}{c|}{\textbf{Supervision}} & \multicolumn{8}{c}{\textbf{Metrics}}  \\ 
        CG & FG & IF & VQA & EM@1 & BLEU-1 & BLEU-2 & BLEU-3 & BLEU-4 & ROUGE & Meteor & CIDEr  \\ \hline
         &  &  & \checkmark & 22.33  & 27.06  & 18.17  & 14.99  & 0.00  & 30.95  & 12.17  & 58.44   \\ 
        \checkmark & \checkmark &  & \checkmark & 22.52  & 27.41  & 19.91  & 17.50  & 13.53  & 31.51  & 12.47  & 60.42   \\ 
        \checkmark &  & \checkmark & \checkmark & 22.98  & 28.13  & 19.16  & 15.61  & 9.91  & 31.99  & 12.50  & 60.62   \\ 
         & \checkmark & \checkmark & \checkmark & 22.02  & 28.37  & 20.23  & 15.75  & 10.88  & 31.28  & 12.44  & 59.44   \\ 
        \checkmark & \checkmark & \checkmark & \checkmark & 22.95  & 28.21  & 20.85  & 17.19  & 13.34  & 32.21  & 12.66  & 61.64  \\ \hline
    \end{tabular}
    \label{tab:abl}
\end{table*}

\subsection{Ablation Study}

To investigate the effect of each module in HSM, we conduct an ablation study by performing experiments in which some intermediate steps are left unsupervised. The results of the experiments are shown in Table \ref{tab:abl}. CG, FG, IF, and VQA represent coarse grounding, fine grounding, inference, and VQA heads, respectively. The supervised heads are marked with a \checkmark.

The results demonstrate that models with less supervision generally perform between answer-centric and full supervision. Since the supervision on each step in the reasoning pathway serves as a checkpoint, reducing supervisions enables the development of more shortcuts, resulting in a deterioration in performance.

\begin{table*}[t]
    \centering
    \caption{Comparison Results on ScanQA's ``Test w/ obj'' Split}
    \begin{tabular}{c|cccccccc}
    \hline
        Model & EM@1 & BLEU-1 & BLEU-2 & BLEU-3 & BLEU-4 & ROUGE & Meteor & CIDEr  \\ \hline
        ScanQA \cite{ScanQA} & 23.45 & 31.56 & 21.39 & 15.87 & 12.04 & 34.34 & 13.55 & 67.29   \\
        CLIP-Guided \cite{CLIPG} & 23.92 & 32.72  & - & - &  14.64  & 35.15  & 13.94  & 69.53   \\ 
        Multi-CLIP \cite{MCLIP} & 24.02 & 32.63  & - & - & 12.65  & 35.46  & 13.97  & 68.70   \\ 
        3DVLP \cite{3DVLP} & 24.58 & 33.15  & 22.65  & 16.38  & 11.23  & 35.97  & 14.16  & 70.18   \\ 
        3D-VisTA \cite{3DVisTA} & 25.94  & 33.07  & 23.36 & 18.16  & 12.89  & 37.21  & 14.75  & 73.39   \\ 
        \textbf{HCNQA (ours)} & \textbf{27.01} & \textbf{33.27} & \textbf{23.69} & \textbf{18.80} & \textbf{14.71} & \textbf{38.46} & \textbf{15.18} & \textbf{75.66}  \\ \hline
    \end{tabular}
    \label{tab:comp}
\end{table*}

\subsection{Comparative Experiments}

In this section, we conduct an experiment to evaluate the performance of our HCNQA model on unseen cases and compare it with previous classification-based models. The results are presented in Table \ref{tab:comp}. The best performance under each metric is highlighted in bold.

The results illustrate that our model outperforms other models on all metrics. This indicates that hierarchical concentration narrowing supervision can guide the model to develop a logical and rigorous intermediate reasoning pathway, thereby enhancing the model's performance and generalizability.

\subsection{Shortcut Measurement}

To directly evaluate our method's capability in suppressing superficial shortcuts, we adopt the perturbation strategy proposed by Ye and Kovashka \cite{shortcut_eval}. Specifically, we generate question variations through synonym substitution, where the performance (EM@1 accuracy) drop magnitude indicates the degree of superficial shortcuts between questions and answers. As shown in Table \ref{tab:sbe}, our method exhibits only $51\%$ of the performance degradation observed in the baseline model, demonstrating its effectiveness in mitigating superficial shortcuts.

\begin{table*}[t]
    \centering
    \caption{Results of Shortcut Behavior Evaluation}
    \begin{tabular}{cccc}
    \hline
        \textbf{Model} & \textbf{~Before Perturbation~} & \textbf{After Perturbation~} & \textbf{Degradation} \\ \hline
        3D-VisTA \cite{3DVisTA} & 23.25 & 22.78 & 0.47 \\ 
        HCNQA (ours) & 23.57 & 23.33 & 0.24 \\ \hline
    \end{tabular}
    \label{tab:sbe}
\end{table*}

\begin{figure}[!ht]
    \centering
    \includegraphics[width=0.98\textwidth]{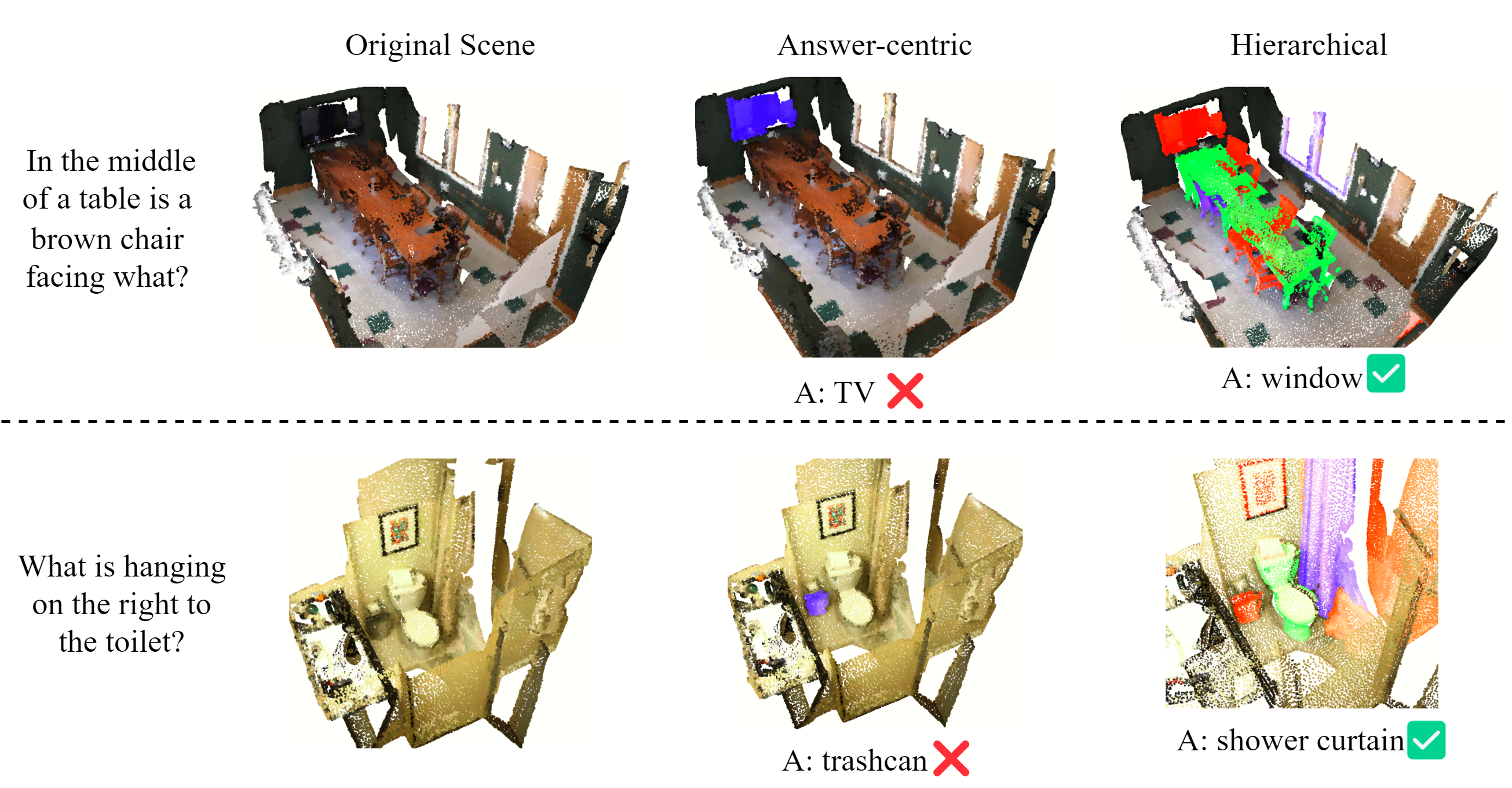}
    \caption{Quantitative results. In the middle column, objects predicted to be in object\_ids by the answer-centric model 3D-VisTA are marked in blue. The right column shows the predictions of our model: all masked objects belong to BoI, green and purple objects are part of OoI, and purple objects also belong to OoT. In the second case, the observer stands in the middle of the room to determine spatial relations by default.}
    \label{fig:qr}
\end{figure}

\subsection{Qualitative Results}

Finally, we demonstrate the superiority of hierarchical concentration narrowing supervision over answer-centric supervision by visualizing representative examples, as shown in Fig. \ref{fig:qr}. The results suggest that our model can generate better answers through a more rigorous reasoning process.

For instance, when responding to the first query, the model must infer multiple spatial relationships of objects (i.e., identifying the brown chair in the middle of the table and the object opposite to the chair). Due to the absence of supervision over the reasoning process in answer-centric models, they overlook the importance of selecting the correct chair as an intermediate step, leading to choosing the wrong chair and ultimately concluding that the chair is facing a TV rather than a window. In contrast, HCNQA, which requires the provision of results for each key intermediate step, can conduct rigorous reasoning throughout the reasoning process, thereby forming a correct response.

\subsection{Discussions}

While effective for multi-hop reasoning, the hierarchical framework underperforms on questions with short or implicit reasoning chains, as its multi-stage mechanism may add unnecessary complexity and propagate errors. For instance, in cases only requiring object detection (e.g., locating a cabinet with the word ``safe''), hierarchical refinement offers little benefit and may mislead the model due to mispredictions in early stages. Potential improvements could include enhancing models' ability in cross-modal perception and reasoning.

\section{Conclusion}

3D-VQA is crucial for enabling models to perceive and reason about the physical world through natural language. In this paper, we propose HCNQA, a 3D-VQA model with a reasoning pathway regulated through hierarchical supervision of the reasoning phases. Extensive experiments demonstrate that our hierarchical concentration narrowing supervision on reasoning pathways can effectively suppress superficial shortcuts, achieving better performance and generalizability.

\subsubsection{\ackname} This study was funded
by China National University Student Innovation and Entrepreneurship Development Program (No. 202414325005).

\bibliographystyle{splncs04}
\bibliography{references}

\end{document}